%% file: cvpr.tex
\documentclass[final]{cvpr}

\usepackage{times}
\usepackage{epsfig}
\usepackage{graphicx}
\usepackage{amsmath}
\usepackage{amssymb}

\usepackage{caption}
\usepackage{subcaption}
\usepackage{comment}
\usepackage{pifont}

\usepackage[pagebackref=true,breaklinks=true,colorlinks,bookmarks=false]{hyperref}

\def\cvprPaperID{5483} 
\def\confYear{CVPR 2021}

\begin{document}

\iftrue
\definecolor{mncolor}{RGB}{255,50,00}
\newcommand\MATTHIAS[1] {\textbf{\textcolor{mncolor}{MN: #1}}}
\definecolor{jtcolor}{RGB}{0,0,255}
\newcommand\JT[1] {\emph{\textcolor{jtcolor}{JT: #1}}}
\definecolor{mzcolor}{RGB}{10,120,10}
\newcommand\MZ[1] {\emph{\textcolor{mzcolor}{MZ: #1}}}
\definecolor{todocolor}{RGB}{255,0,00}
\newcommand\TODO[1] {\emph{\textcolor{todocolor}{TODO: #1}}}
\newcommand\GG[1] {\emph{\textcolor{todocolor}{GG: #1}}}

\else 
\newcommand\MATTHIAS[1] {}
\newcommand\JT[1] {}
\newcommand\TODO[1] {}
\newcommand\GG[1] {}

\fi 
\newcommand{\norm}[1]{\left\lVert#1\right\rVert}
\newcommand{\cmark}{\textcolor{mzcolor}{\ding{51}}}  %
\newcommand{\xmark}{\textcolor{mncolor}{\ding{55}}}%

\title{Dynamic Neural Radiance Fields for \\ Monocular 4D Facial Avatar Reconstruction}

\author{
Guy Gafni$^1$~~~
Justus Thies$^1$~~~
Michael Zollh\"{o}fer$^2$~~~
Matthias Nie\ss{}ner$^1$
\\ 
$^1$Technical University of Munich~~~
$^2$Facebook Reality Labs
}



\twocolumn[{
	\renewcommand\twocolumn[1][]{#1}%
	\maketitle
	\begin{center}
		\includegraphics[width=1\textwidth]{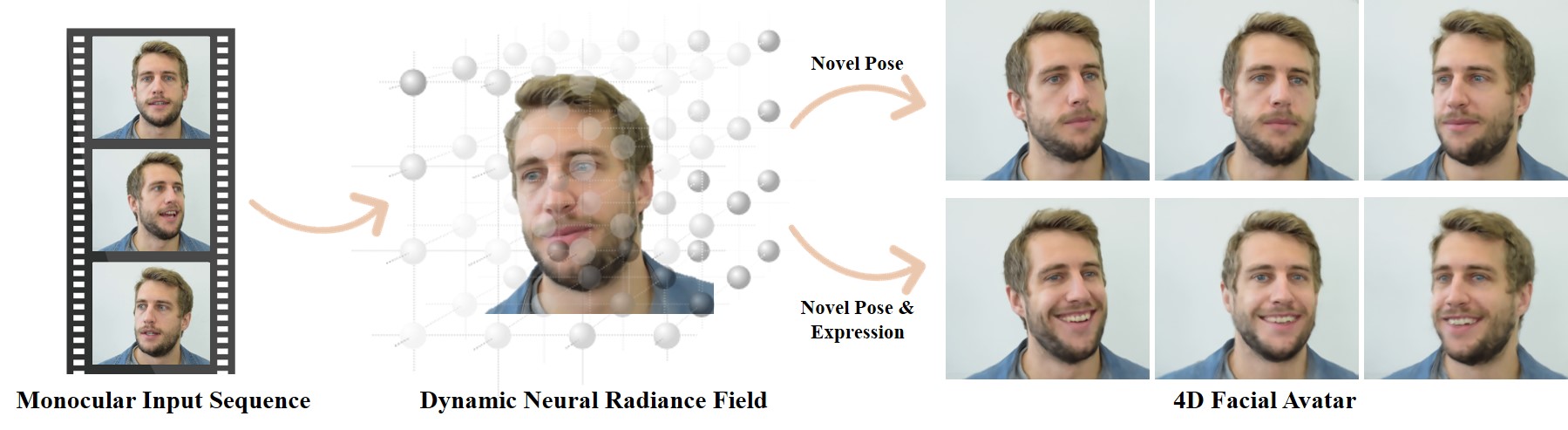}
		\captionof{figure}{Given a monocular portrait video sequence of a person, we reconstruct a dynamic neural radiance field representing a 4D facial avatar, which allows us to synthesize novel head poses as well as changes in facial expressions.}
		\label{fig:teaser}
	\end{center}
	\vspace{\baselineskip}
	\vspace{-0.1cm}
}]

\begin{abstract}

\input{latex/chapters/00_abstract}
\end{abstract}

\section{Introduction}
\input{latex/chapters/01_introduction}

\section{Related Work}
    \input{latex/chapters/02_related}
\section{Method}
    \input{latex/chapters/03_method}

\section{Results}

\input{latex/chapters/04_results}

\section{Limitations}
\input{latex/chapters/05_limitations}
\section{Conclusion}
\input{latex/chapters/06_discussion}
\section*{Acknowledgments}
    \input{latex/chapters/08_acknowledgements}

{\small
\bibliographystyle{ieee_fullname}
\bibliography{egbib}
}
\newpage
\appendix

\input{latex/chapters/07_appendix}

\end{document}

%% file: latex/chapters/00_abstract.tex
We present dynamic neural radiance fields for modeling the appearance and dynamics of a human face\footnote{\url{gafniguy.github.io/4D-Facial-Avatars}}.
Digitally modeling and reconstructing a talking human is a key building-block for a variety of applications. %
Especially, for telepresence applications in AR or VR, a faithful reproduction of the appearance including novel viewpoint or head-poses is required.
In contrast to state-of-the-art approaches that model the geometry and material properties explicitly, or are purely image-based, we introduce an implicit representation of the head based on scene representation networks.
To handle the dynamics of the face, we combine our scene representation network with a low-dimensional morphable model which provides explicit control over pose and expressions.
We use volumetric rendering to generate images from this hybrid representation and demonstrate that such a dynamic neural scene representation can be learned from monocular input data only, without the need of a specialized capture setup.
In our experiments, we show that this learned volumetric representation allows for photo-realistic image generation that surpasses the quality of state-of-the-art video-based reenactment methods.
\vspace{-0.1cm}

%% file: latex/chapters/01_introduction.tex
Reconstructing 4D models of humans and, especially, the human face, is an ongoing research problem in the field of computer vision and computer graphics.
4D avatars are essential for augmented reality (AR) and virtual reality (VR) telepresence applications as well as for video editing, such as visual dubbing in movie productions.
These applications need a faithful reconstruction of the human's appearance, as well as the ability to change the viewpoint or head pose (especially, in VR) and the expressions (e.g., for visual dubbing).
Representing a human head with explicit geometry and material properties (e.g., albedo, reflectance) is challenging; the skin has effects like subsurface scattering, the eyes are highly reflective and the hair has a complex geometry with fine scale details.
While the explicit reconstruction of high quality geometry of the skin surface in a multi-view studio setup is tractable~\cite{Bee11,Gotardo2018,zollhoefer2018facestar}, hair is often approximated by retrieval and refinement of hair styles~\cite{hu2015single,zhou2018hair}, which leads to an unrealistic visual reproduction.

To handle the material properties and complex geometry of a 4D facial avatar, we introduce \textit{dynamic neural radiance fields}.
Our approach is a neural rendering method combining classical volume rendering with a novel neural scene representation network to achieve novel head pose and expression synthesis.
In contrast to related work on learned scene representations that focuses on static objects and multi-view input data, we are able to represent the dynamically changing surface of a human's face only based on \emph{monocular} camera recordings.
The representation is a stepping stone towards reconstruction of 4D facial avatars using commodity hardware, allowing for novel viewpoint synthesis of the head in a virtual reality setting, pose changes in videos or even facial reenactment where the expressions of one person are transferred to another person (represented by our scene representation network).
The learned scene representation is a volumetric representation which is key to capturing hair, but also the mouth interior where classical methods struggle because of missing 3D geometry.
The implicit representation of the geometry and appearance defines a continuous function in space that does not suffer from discretization artifacts of voxel grids (e.g., limited resolution) and is optimized to represent the head as good as possible w.r.t.~the final re-renderings and the underlying network architecture.
In contrast to state-of-the-art facial reenactment and video editing approaches~\cite{kim2018deep,thies2019}, our volumetric approach is able to synthesize 3D-consistent content with large head pose changes.
Large head pose changes (or view changes) are required for VR or AR applications, but can also be used for face frontalization or to dampen the variance of motion.
The semantically meaningful conditioning used in our method also allows for user-driven edits of a video in a post-processing scenario.
Specifically, our method is based on a short portrait video sequence of a person.
To represent the expressions of the face, we leverage a low-dimensional morphable model~\cite{Blanz1999,thies2016face}.
Given the pose of the model and the expression parameters of a specific frame of a sequence that has to be synthesized, we dispatch rays in a canonical space where our neural scene representation network is embedded.
Along the rays, we perform volumetric integration of density and color values predicted by our scene representation network that is inspired by the work of Mildenhall \etal.~\cite{mildenhall2020nerf}, which focuses on high quality multi-view reconstruction of a static scene.
Note that the scene representation network is not only conditioned on the sample point locations but also on the expressions of the morphable model which allows for the dynamically changing content that has to be stored in the neural network.
During test time, this conditioning allows us to apply novel head poses as well as expressions to synthesize a new image.
We demonstrate that our technique is able to faithfully represent a 4D facial avatar and show photo-realistic results that surpass state-of-the-art facial reenactment methods.

\noindent
To summarize, we show that neural scene representation networks can be used to store and represent the dynamically changing surface of a human head in a controllable manner.
Our contributions are:
\begin{itemize}
    \item Dynamic Neural Radiance Fields to represent 4D facial avatars based on a low dimensional morphable model.
    \item An efficient end-to-end learnable approach that uses a single camera to reconstruct such a radiance field.
\end{itemize}

%% file: latex/chapters/02_related.tex
Our approach is a neural rendering method to represent and generate images of a human head.
It is related to recent approaches on neural scene representation networks, as well as neural rendering methods for human portrait video synthesis and facial avatar reconstruction.
In the following, we discuss the most related literature in the two fields in detail.

\paragraph{Face Reconstruction based on a Morphable Model}
For a summary of facial reconstruction methods, we refer to the state-of-the-art report of Zollh\"ofer \etal~\cite{zollhoefer2018facestar}.
Our method is built upon a low-dimensional morphable model~\cite{Blanz1999,thies2016face} which is a building block of numerous facial reconstruction and animation approaches~\cite{Garrido2014,Garrido2015,Weise2009,Weise2011,Thies15,thies2016face, Blanz2003,Blanz2004,kim2018deep,thies2019,thies2020nvp}.
In contrast to these methods, we are not relying on the coarse representation of the surface of the face.
Some methods \cite{Chen:2013,Li2013,Bouaziz2013,hsieh2015unconstrained} also focused on corrective shapes \cite{Bouaziz2013}, dynamically adapting the blendshape basis \cite{Li2013} or applied non-rigid mesh deformation \cite{Chen:2013} to compensate for the coarse geometry of the morphable model.
In our approach, we are not relying on a template mesh or an explicit surface representation.
Instead, we represent the geometry and appearance implicitly using a deep neural network and use volumetric rendering to generate new images.

\paragraph{Human Avatar Reconstruction}

The goal of our approach is the photo-realistic reproduction of the head of a human observed from a monocular input stream.
Multiple methods exist that reconstruct personalized face rigs based on hand-held monocular input.
Ichim \etal~\cite{Ichim2015} assume a static pose and expression to reconstruct the head via multi-view stereo.
Hu \etal~\cite{Hu2017} combine face digitization and hair reconstruction to estimate the head geometry and appearance from a single image.
Our implicit function represents the face region as well as the hair in a single formulation, also recovering the volumetric effects of the hair.

\begin{figure*}[t]
    \centering
	\includegraphics[width=\linewidth]{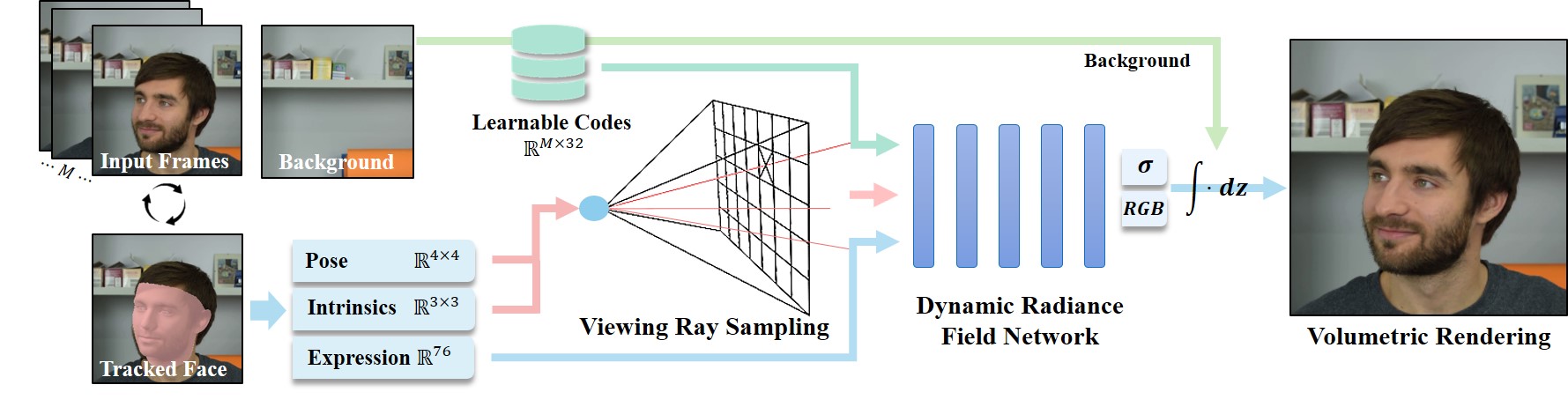}
	\caption{
	    Overview of our 4D facial avatar reconstruction pipeline.
	    Given a portrait video and an image without the person (background image) as input, we apply facial expression tracking using a 3D morphable model. Based on the estimated pose and expression, we use volumetric rendering to synthesize the image of the face. The samples along the viewing rays are input to our dynamic radiance field, which is additionally conditioned on a learnable per-frame latent code. Since the background is static, we set the color of the last sample point of each ray to the corresponding value of the background image.
    }
	\label{fig:pipeline_training}
\end{figure*}

\paragraph{Human Portrait Video Synthesis}
There is a wide range of human portrait video synthesis and editing approaches.
Classical computer graphics approaches use a morphable model reconstruction and forward rendering with optimized textures and a texture atlas for different mouth interiors (since the morphable model is too coarse to model the mouth cavity)~\cite{Garrido2015,Garrido2014,thies2016face,thies2018headon,thies2018facevr}.
Image warping is used in Averbuch-Elor \etal~\cite{averbuch2017}.
In contrast, the most recent approaches are hybrids between classical rendering and learned image synthesis.
Deep Video Portraits~\cite{kim2018deep} is one of the first methods that uses rendered correspondence maps together with an image-to-image translation network to output photo-realistic imagery.
Deferred Neural Rendering~\cite{thies2019,thies2020nvp} extends this idea, by introducing neural feature descriptors that are embedded on the surface of a coarse reconstructed face mesh.
Instead of this dense conditioning input or rendered feature maps, there are also methods that work on rendered facial landmarks~\cite{Zakharov_2019_ICCV,Chen_2020_CVPR,Xu_2020_CVPR}.
These approaches can also be applied to single images.
First Order Motion Model~\cite{Siarohin_2019_NeurIPS} is a data-driven approach that decouples appearance and motion in a video of a specific class (e.g., human faces) and allows application of the motion in a source video to a target image.

\paragraph{Neural Scene Representation Networks}
Neural scene representation networks are building blocks of several neural rendering and neural reconstruction approaches.
A summary of neural rendering approaches is given in the state-of-the-art report of Tewari \etal~\cite{tewari2020neuralrendering}.
Sitzmann \etal~\cite{sitzmann2019srns} introduced neural scene representation networks (SRNs).
The geometry and appearance of an object is represented as a neural network that can be sampled at points in space.
A ray marching approach is used to sample from the neural network to render the reconstructed surface.
On synthetic data, they show the capabilities of such an implicit representation. 
A neural scene representation network is a compact representation that does not suffer from limited resolution as for example, discrete grid structures that store learnable features, e.g., Deep Voxels~\cite{Sitzmann:2018:DeepVoxels} or Neural Volumes~\cite{Lombardi2019}.
Mildenhall \etal\cite{mildenhall2020nerf} extend this idea to store radiance fields in a neural network.
They assume a static object and multi-view data.
A key contribution is the volumetric integration and the usage of positional encoding for higher detailed reconstructions.
Follow-up work extends this idea by using different positional encodings~\cite{tancik2020fourier} and in-the-wild training data including appearance interpolation~\cite{martinbrualla2020nerfw}.
Concurrent work of Sitzmann \etal~\cite{sitzmann2019siren} proposes the usage of sinusoidal activation functions for the scene representation network.
Neural Sparse Voxel Fields \cite{liu2020neural} employ an Octree to cull empty space and speed up rendering.
While these methods have a focus on static objects, we are dealing with a dynamically changing surface of a face.
We use a similar volumetric integration scheme to \cite{mildenhall2020nerf} with an additional layer for the static background.
The dynamic neural scene representation is not only conditioned on the sample position and view direction, but also on the facial deformations.

%% file: latex/chapters/03_method.tex
\begin{figure}[htbp]
    \centering
    \includegraphics[width=0.9\linewidth]{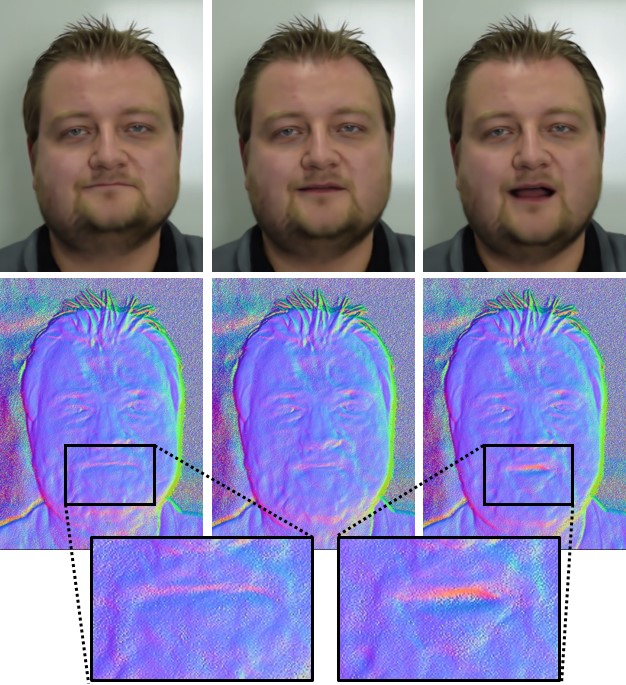}
    \caption{
    Our dynamic radiance field allows for manual editing via the expression vector $\delta$.
    In the middle we show the reconstruction of the original expression. On the left and right we show the results of modifying the blendshape coefficient of the mouth opening (left $-0.4$, right $+0.4$).
    The bottom row shows the corresponding normal maps computed via the predicted depth.
    As can be seen, the dynamic radiance field adapts not only the appearance, but also the geometry according to the input expression.
    }
    \label{fig:geometry}
\end{figure}

Our approach enables 4D reconstruction of a facial avatar based on a single portrait video of a person (see Fig.~\ref{fig:pipeline_training}).
The geometry and appearance of the human head is represented implicitly by a neural scene representation network.
Specifically, the neural scene representation network stores a dynamic neural radiance field which is used during volumetric rendering.
The dynamics of the human face, i.e., the facial expressions, are first captured with a state-of-the-art face tracking approach~\cite{thies2016face}.
The resulting low dimensional expression parameters of the morphable model are used as conditioning for the neural scene representation network.
Note that the expression parameters have semantic meaning allowing us to change specific expressions (see Fig.~\ref{fig:geometry}) or to apply the expressions of a different recorded person~(see Fig.~\ref{fig:reenactment}). 
In addition, we employ the pose parameters (rotation, translation) of the face tracking to transform the rays into a canonical space that is shared by all frames.

\subsection{Dynamic Neural Radiance Fields}
We represent the dynamic radiance field of a talking human head using a multi-layer perceptron (MLP) $\mathcal{D}_{\theta}$ that is embedded in a canonical space.
As the dynamic radiance field is a function of position $\mathbf{p}$, view $\vec{v}$ and dynamics in terms of facial expressions $\mathbf{\delta}$, we provide these inputs to the MLP which outputs color as well as density values for volumetric rendering:
\begin{equation}
    \mathcal{D}_{\theta}(\mathbf{p}, \vec{v}, \mathbf{\delta}, \mathbf{\gamma}) = (RGB, \sigma)
\end{equation}
Note, to compensate for errors in the facial expression and pose estimation, we also provide a per-frame learnable latent code $\mathbf{\gamma}$ to the MLP.
Instead of directly inputting the canonical position $\mathbf{p}$ and viewing direction $\vec{v}$, we use positional encoding as introduced by Mildenhall \etal.~\cite{mildenhall2020nerf}.
In our experiments, we use $10$ frequencies for the position $\mathbf{p}$ and $4$ frequencies for the viewing direction $\vec{v}$.

\paragraph{Dynamics Conditioning}
A key component of the dynamic neural radiance fields is the conditioning on the dynamically changing facial expressions.
The facial expressions $\mathbf{\delta}$ are represented by coefficients of a low dimensional delta-blendshape basis of a morphable model ($\mathbf{\delta} \in \mathbb{R}^{76}$).
To estimate the per-frame expressions $\mathbf{\delta}_i$, we use an optimization-based facial reconstruction and tracking pipeline~\cite{thies2016face}.
Note that these expression vectors only model the coarse geometric surface changes and do not model changes of for example the eye orientation.
Besides expression parameters, we also store the rigid pose $P_i \in \mathbb{R}^{4\times4}$ of the face which allows us to transform camera space points to points in the canonical space of the head. 

\begin{figure*}[htbp]
    \centering
    \includegraphics[width=0.9\linewidth]{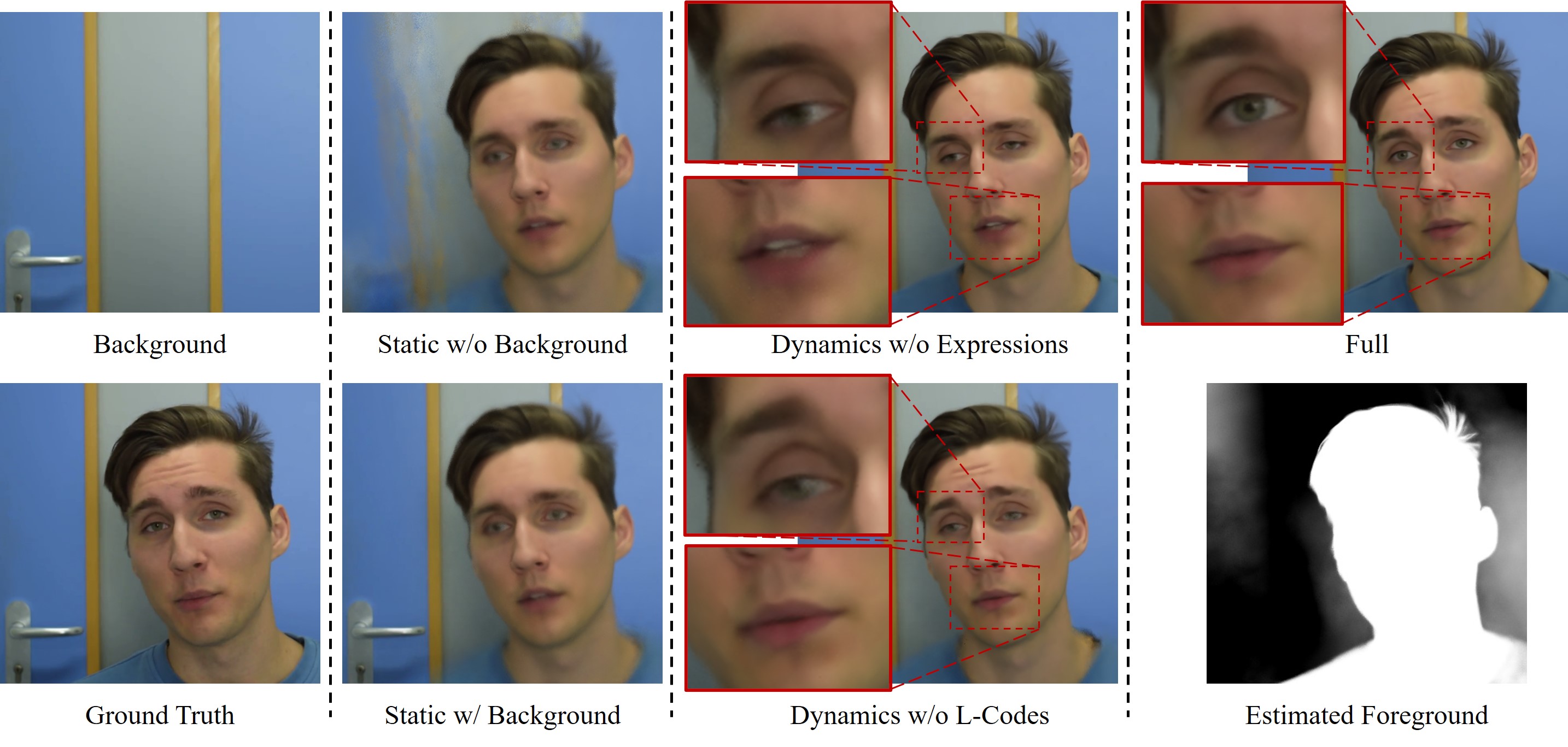}
    \caption{
    The background image enables us to faithfully reproduce the entire image. While the dynamics are mainly conditioned on the facial expressions, the learnable latent codes improve the sharpness of the image significantly.
    Our method also implicitly gives access to a foreground segmentation.
    Note that the shown images are from the test set (latent code is taken from the first frame of training set).
    }
    \label{fig:latent_codes}
\end{figure*}

To compensate for missing information of the expression vectors, we introduce learnable latent codes $\mathbf{\gamma}_i$ (one for each frame).
In the experiments, we are using $\mathbf{\gamma}_i \in \mathbb{R}^{32}$ and regularize them via an $\ell_2$ loss using weight decay ($\lambda=0.05$).
In Fig.~\ref{fig:latent_codes}, we show that the latent code improves the overall sharpness of the reconstruction.
Evaluating the Learned Perceptual Image Patch Similarity (LPIPS)~\cite{zhang2018perceptual} metric for the generated images with and without latent codes results in 0.059 and 0.068, respectively.

\subsection{Volumetric Rendering of Portrait Videos}
In our experiments, we assume a static camera, and a static background.
The moving and talking human in the training portrait video is represented with a dynamic neural radiance field as introduced in the previous section.
To render images of this implicit geometry and appearance representation, we use volumetric rendering.
We cast rays through each individual pixel of a frame, and accumulate the sampled density and RGB values along the rays to compute the final output color.
Using the tracking information $P$ of the morphable model, we transform the ray sample points to the canonical space of the head model and evaluate the dynamic neural radiance field at these locations.
Note that this transformation matrix $P$ gives us the control over the head pose during test time.
We use a similar two-stage volumetric integration approach to Mildenhall \etal.~\cite{mildenhall2020nerf}.
Specifically, we have two instances of the dynamic neural radiance field network, a coarse and a fine one.
The densities predicted by the coarse network are used for importance sampling of the query points for the fine network, such that areas of high density are sampled more.
The expected color $\mathcal{C}$ of a camera ray $\mathbf{r}(t)=\mathbf{c}+t\vec{d}$ with camera center  $\mathbf{c}$, viewing direction $\vec{d}$ and near $z_\text{near}$ and far bounds $z_\text{far}$ is evaluated as:
\begin{equation}
    \label{eq:integration}
    \mathcal{C}(\mathbf{r};\theta,P,\mathbf{\delta},\mathbf{\gamma})=\int_{z_\text{near}}^{z_\text{far}}{\sigma_{\theta}\left(\mathbf{r}\left(t\right)\right)\cdot\text{RGB}_{\theta}\bigl(\mathbf{r}\left(t\right),\vec{d}\bigr)}\cdot T(t)dt,
\end{equation}
where $RGB_{\theta}(\cdot)$ and $\sigma_{\theta}(\cdot)$ are computed via the neural scene representation network $\mathcal{D}_{\theta}$ at a certain point on the ray with head pose $P$, expressions $\mathbf{\delta}$ and learnable latent code $\mathbf{\gamma}$.  $T(t)$ is the accumulated transmittance along the ray from $z_\text{near}$ to $t$:
\begin{equation}
    T(t) = \exp{  \left( -\int_{ z_\text{near} }^{t}{\sigma_{\theta}\left(\mathbf{r}\left(s\right)\right)ds}\right) . }
\end{equation}
Note that the expected color is evaluated for both the coarse and the fine networks (with learnable weights $\theta_{coarse}$ and $\theta_{fine}$, respectively) to compute corresponding reconstruction losses at train time (see Eq.~\ref{eq:train_loss}).
We decouple the static background and the dynamically changing foreground by leveraging a single capture of the background $\mathcal{B}$ (i.e., without the person).
The last sample on the ray $\mathbf{r}$ is assumed to lie on the background with a fixed color, namely, the color of the pixel corresponding to the ray, from the background image.
Since the volumetric rendering is fully differentiable, the network picks up on this signal,
and learns to predict low density values for the foreground samples if the ray is passing through a background pixel, and vice versa - for foreground pixels, i.e., pixels that correspond to torso and head geometry, the networks predict higher densities, effectively ignoring the background image.
This way, the network learns a foreground-background decomposition in a self-supervised manner (see Fig.~\ref{fig:latent_codes}).

\subsection{Network Architecture and Training}

As mentioned above, the dynamic neural radiance field is represented as an MLP.
Specifically, we use a backbone of $8$ fully-connected layers, each $256$ neurons-wide, followed by ReLu activation functions.
Past the backbone, the activations are fed through a single layer to predict the density value, as well as a $4$-layer, $128$ neuron-wide branch to predict the final color value of the query point.
We optimize the network weights of both the coarse and the fine networks based on a photo-metric reconstruction error metric over the training images $I_i$ ($i \in [1,M]$):
\begin{equation}
    \label{eq:train_loss}
    L_{total} = \sum_{i=1}^{M} L_i(\theta_{coarse}) + L_i(\theta_{fine})
\end{equation}
with
\begin{equation}
    \label{eq:train_loss2}
    L_{i}(\theta) = \sum_{j \in \text{pixels}}{ \norm{ \mathcal{C}\bigl(\textbf{r}_j;\theta,P_i,\mathbf{\delta}_i,\mathbf{\gamma}_i) - I_i[j]}^2 } .
\end{equation}
For each training image $I_i$ and training iteration, we sample a batch of $2048$ viewing rays through the image pixels.
We use a bounding box of the head (given by the morphable model) to sample the rays such that 95\% of them correspond to pixels within the bounding box and, thus allowing us to reconstruct the face with a high fidelity.
Stratified sampling is used to sample $64$ points along each ray, which are fed into the coarse network $\mathcal{D}_{\theta_{coarse}}$.
Based on the density distribution along the ray, we re-sample $64$ points and evaluate the color integration (see Eq.~\ref{eq:integration}) using the fine network $\mathcal{D}_{\theta_{fine}}$.
Our method is implemented in PyTorch~\cite{pytorch}.
Both networks and the learnable codes $\mathbf{\gamma}_i$ are optimized using Adam~\cite{adam} ($lr=0.0005$).
In our experiments, we use $512\times512$ images and train each model for $400k$ iterations.

\begin{figure*}[]
    \includegraphics[width=1.0\linewidth]{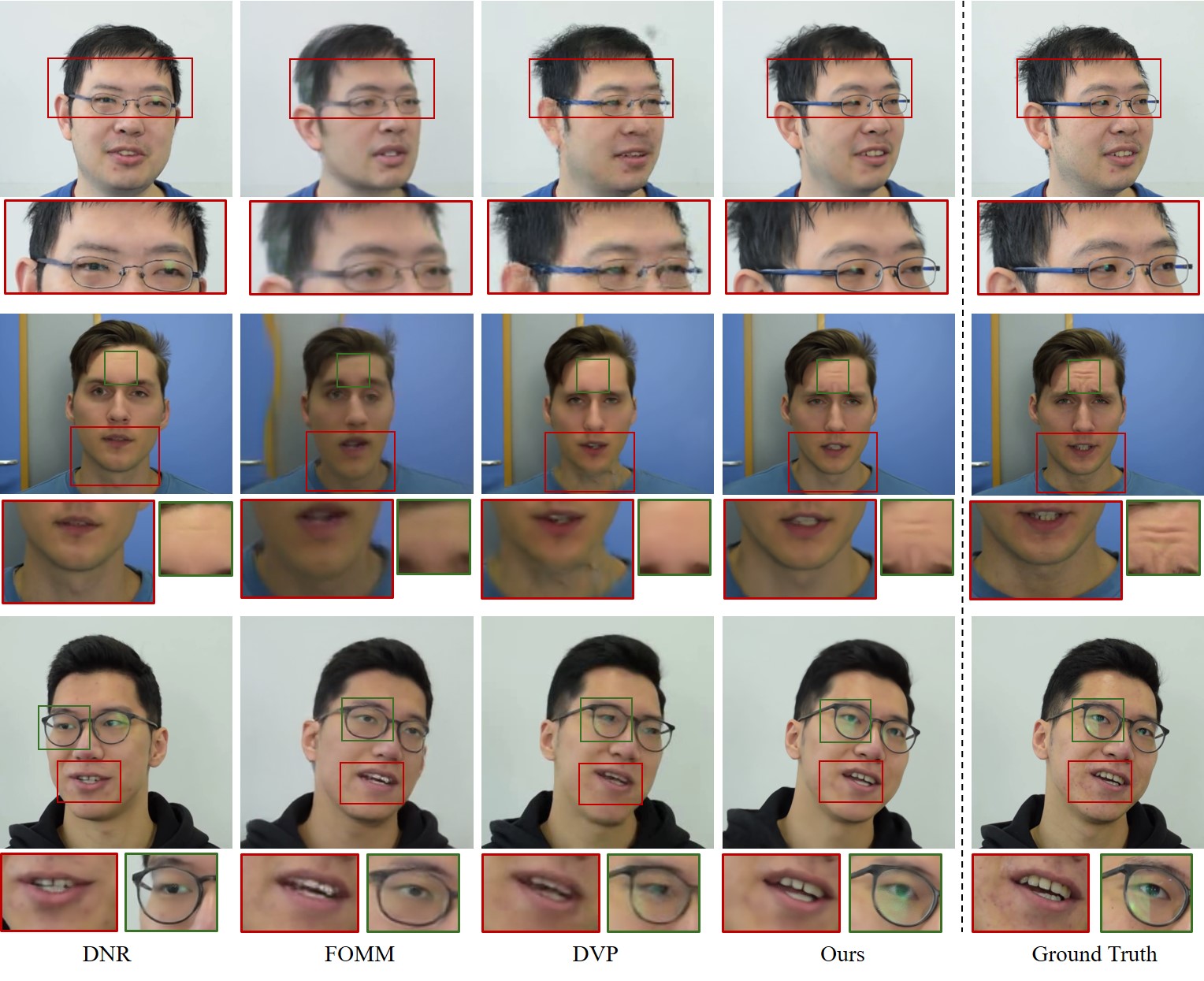}
    \caption{Comparison to state-of-the-art facial reenactment methods based on self-reenactment. From left to right: Deferred Neural Rendering (DNR)~\cite{thies2019}, First Order Motion Models (FOMM)~\cite{Siarohin_2019_NeurIPS}, Deep Video Portraits (DVP)~\cite{kim2018deep}, Ours and the ground truth image.
    Note that DNR does not provide control over the pose parameters and only changes the facial expressions.
    As can be seen, our approach faithfully reconstructs the expression and appearance of the faces, and can also represent the geometry of the glasses including the view-dependent effects (see last row). 
    }%
    \label{fig:results_compare}%
\end{figure*}

%% file: latex/chapters/04_results.tex
Our approach allows the reconstruction of a 4D facial avatar based on monocular video sequences (see Sec.~\ref{sec:main_results}).
In the following, we analyze our method qualitatively and quantitatively on real data (Sec.~\ref{sec:training_data}).
Specifically, we show comparisons to state-of-the-art facial reenactment methods (Sec.~\ref{sec:comp}) and discuss the conducted ablation studies of our method (Sec.~\ref{sec:ablation}).
The advantages of our approach can best be seen in the supplemental video, especially, the 3D consistency of pose changes and the faithful reproduction of the appearance.
\subsection{Monocular Training Data}
\label{sec:training_data}
Our method uses short monocular RGB video sequences.
We captured various human subjects with a Nikon D5300 DSLR camera at a resolution of $1920 \times 1080$ pixels with a framerate of $50$ frames per second.
The images are cropped to $1080 \times 1080$ and scaled to $512 \times 512$.
The sequences have a length of about $2$ min ($6000$ frames).
We hold out the last 20 seconds ($1000$ frames) to serve as a test sequence for each reconstruction.
The subjects were asked to engage in normal conversation, including expressions like smiling as well as head rotations.

\begin{figure*}[h!]
    \centering
    \includegraphics[width=0.9\textwidth]{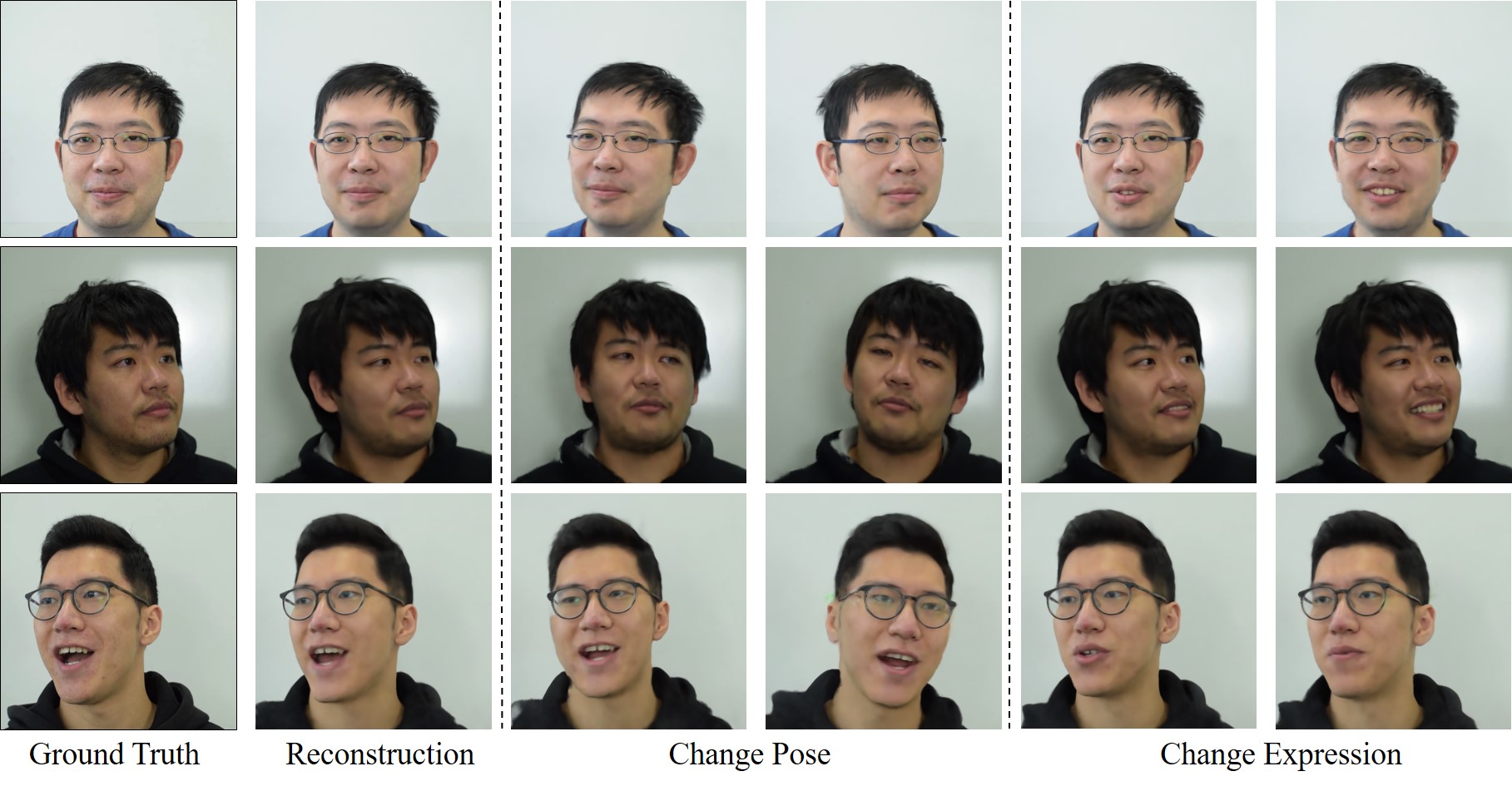}
    \caption{We demonstrate the manual controllability of pose and expression using our 4D facial avatars reconstructed from monocular video inputs.
    Specifically, we demonstrate 3D consistent novel head pose synthesis and expression changes (by changing the `open mouth' blendshape coefficient).
    }
    \label{fig:main_results}
\end{figure*}

\subsection{Comparison to the State of the Art}
\label{sec:comp}

From the application stand-point, our method competes with state-of-the-art facial reenactment methods that allow to apply pose and expression changes.
Specifically, we compare our method with Deep Video Portrait of Kim \etal.~\cite{kim2018deep}, Deferred Neural Rendering of Thies \etal.~\cite{thies2019} and First-Order Motion Models of Siarohin \etal.~\cite{Siarohin_2019_NeurIPS}.
%
In Fig.~\ref{fig:results_compare} we show qualitative results of the above-mentioned and our own method in a self-reenactment scenario.
As can be seen, our method is able to reproduce the photo-realistic appearance of the subjects.
In contrast to the other methods, our approach generates 3D consistent results including view-dependent effects like the reflections on the glasses.
Especially, synthesizing new head rotations is challenging for the baseline methods.
Note that the approach of Thies \etal.~\cite{thies2019} only controls the facial expressions and not the pose.
To quantitatively evaluate our method and the other two approaches, we compute the mean $L_1$-distance, Peak Signal-to-Noise Ratio (PSNR), and Structure Similarity Index (SSIM)~\cite{ssim}, as well as the Learned Perceptual Image Patch Similarity (LPIPS)~\cite{zhang2018perceptual} metric.
The results are listed in Tab.~\ref{table:comparison_other_methods}.

\begin{table}
    \begin{center}
    \begin{tabular}{|l|c|c|c|c|}
        \hline
        Method & $L_1 \downarrow$ & PSNR $\uparrow$ & SSIM $\uparrow$ & LPIPS $\downarrow$ \\
        \hline\hline
        FOMM~\cite{Siarohin_2019_NeurIPS} & $0.036$ & $23.77$ & $0.91$ & $0.16$  \\
        DVP~\cite{kim2018deep}       & $0.021$ & $25.67$ & $0.93$ & $0.10$ \\ 
        \hline
        Ours (no BG)                      & $0.035$ & $23.52$ & $0.90$ & $0.18$ \\

        Ours (no dyn.)                    & $0.024$  & $26.65$ & $0.93$ & $0.11$ \\
        Ours (full) & $\mathbf{0.019}$ & $ \mathbf{26.85}$ & $\mathbf{0.95}$ &  $\mathbf{0.06}$ \\
        \hline
    \end{tabular}
    
    \end{center}
    \caption{Quantitative evaluation of our method in comparison to state-of-the-art facial reenactment methods based on self-reenactment (see. Fig.~\ref{fig:results_compare}). Ours (no dyn.) refers to our method without conditioning on dynamics. Ours (no BG) is our method without background image input.}
    \label{table:comparison_other_methods}
\end{table}

\subsection{Novel Pose and Expression Synthesis}
\label{sec:main_results}

\begin{figure*}
    \centering
    \includegraphics[width=0.95\linewidth]{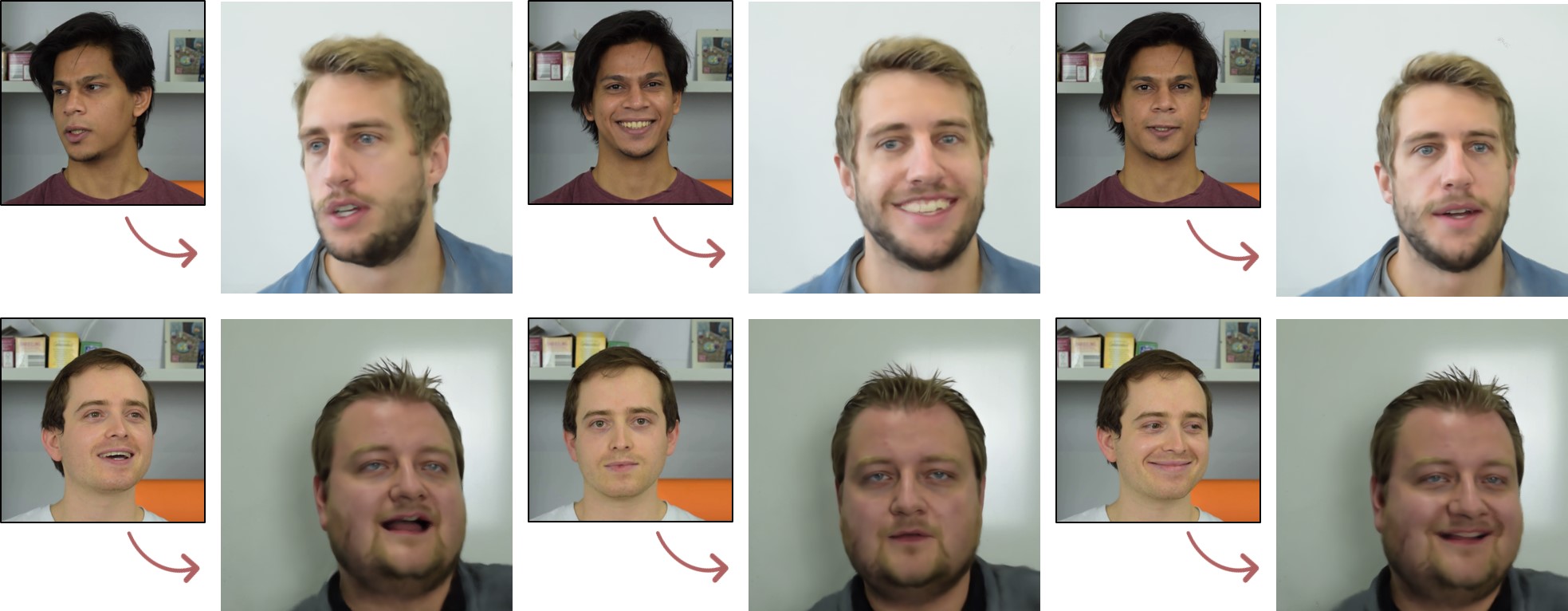}
    \caption{Our 4D facial avatars allow for facial reenactment, where the expressions of a source person are transferred to a target actor which we represent with our dynamic neural radiance field. Note that for facial reenactment we only need to train a model for the target actor; the expressions and pose changes from the source actor can be obtained in real-time~\cite{thies2016face}.}%
    \label{fig:reenactment}
\end{figure*}

The goal of our method is the reconstruction of a 4D facial avatar with explicit control over pose and expressions.
We show several reconstructed avatars in Fig.~\ref{fig:main_results} including synthesized images with modified facial expressions and rigid pose.
The results are best seen in the supplemental video, which shows that our dynamic neural scene representation can effectively store the appearance and geometry of a talking head.
In addition to the manual expression and pose edits, we demonstrate facial reenactment where we transfer the facial expressions of one person to another (see Fig.~\ref{fig:reenactment}).

\subsection{Ablation Studies}
\label{sec:ablation}
Our method assumes a static background and receives a background image as input.
This background image helps to disentangle the foreground (4D facial avatar) and the background (see Fig.~\ref{fig:latent_codes}).
The conditioning on the facial dynamics in form of the per-frame facial expression coefficients and learnable latent codes is one of the key components of our approach.
Note that during test time we always employ the latent code from the first frame of the training set.
Besides qualitative results, we also list a quantitative evaluation in Tab.~\ref{table:comparison_other_methods}.
As can be seen, all components of our approach improve the quality of the results.
While static neural radiance fields can achieve satisfactory quality with as few as 100 posed images~\cite{mildenhall2020nerf}, our method requires more training data.
In our setting the dynamic radiance field is required to generalize over the space of expression vectors.
To quantify the need of a large training corpus, we conducted experiments by only training on the first halves and quarters of the training sequences, such that a lower variety of expressions and poses is seen during training.
The measured degradation in quality as we train with less data is shown in Tab. \ref{table:ablation_data}.

\begin{table}
    \begin{center}
    \begin{tabular}{|l|c|c|c|c|}
    \hline
    Method & $L_1 \downarrow$ & PSNR $\uparrow$ & SSIM $\uparrow$ & LPIPS $\downarrow$ \\
    \hline\hline
    Ours ($25\%$) & $0.029$ & $24.22$ & $0.93$ & $0.09$ \\ 
    Ours ($50\%$)& $0.024$ & $25.47$ & $0.94$ & $0.07$ \\ \hline
    Ours (full) & $\mathbf{0.019}$ & $ \mathbf{26.85}$ & $\mathbf{0.95}$ &  $\mathbf{0.06}$ \\
    \hline
    \end{tabular}
    \end{center}
    \caption{Ablation study w.r.t. training corpus size. All metrics significantly benefit from a larger training corpus.}
    \label{table:ablation_data}
\end{table}

%% file: latex/chapters/05_limitations.tex
In comparison to the state-of-the-art methods, our volumetric 4D representation of the head shows significantly better reconstruction abilities both quantitatively and qualitatively.
Nevertheless, our approach still has limitations which we want to discuss in the following.
The morphable model we use~\cite{Blanz1999,thies2016face} does not model eye blinks and eye movements, thus, these deformations can not explicitly be controlled in our approach.
However, eye blinks are implicitly correlated with other expression parameters, and consequently modelled by our method.
Our method is not restricted to this morphable model and could also be used with more sophisticated models that include these additional control handles.
The focus of our work is the reconstruction of the human head; we are currently not modelling the dynamics of the upper body.
In future work, our approach can be extended to these regions (given a consistent tracking of the torso).

%% file: latex/chapters/06_discussion.tex
We have presented a novel method for learning and rendering controllable 4D facial avatars based on dynamic neural radiance fields.
Using volumetric rendering, we are able to capture arbitrary geometry and topology such as hair, eyeware, hats etc., which typically is not supported by morphable model based methods.
In contrast to other volumetric approaches which require an expensive calibrated multi-view rig, our method requires only a single view from a fixed camera, such as a webcam.
This makes our method suitable for capturing avatars of end users at home, using only 2 minutes of their time. 
The reconstructed avatars can be rendered photo-realistically under novel poses and expressions.
The achieved quality beats state-of-the-art facial reenactment methods both quantitatively and qualitatively.\\

%% file: latex/chapters/08_acknowledgements.tex
This work was supported by a TUM-IAS Rudolf Mößbauer Fellowship, the ERC Starting Grant \textit{Scan2CAD} (804724), the German Research Foundation (DFG) Grant \textit{Making Machine Learning on Static and Dynamic 3D Data Practical}, and a Google Research Grant. We would like to thank Mohamed Elgharib for running Deep Video Portraits~\cite{kim2018deep} on our data, and Angela Dai for the video voice-over. 

%% file: latex/chapters/07_appendix.tex

\twocolumn[{%
	\renewcommand\twocolumn[1][]{#1}%
	\begin{center}
	    \vspace{-0.4cm}
		\includegraphics[width=0.9\linewidth]{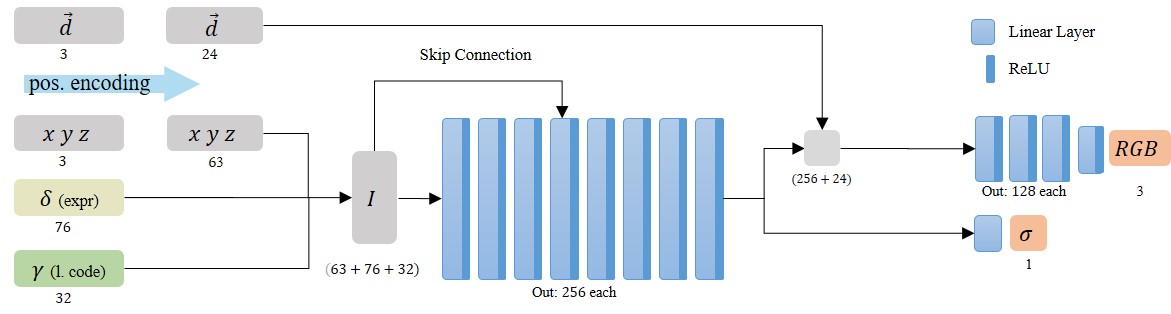}
	\captionof{figure}{
    Our Dynamic Neural Radiance Field is represented as a multi-layer perceptron (MLP).
    	As input it gets the viewing direction $\vec{d}$, the sample position $(x,y,z)$, the expression coefficients $\delta$ as well as the learned latent codes $\gamma$.
    	The viewing direction as well as the position are encoded using positional encoding~\cite{mildenhall2020nerf}.
    	The MLP consists of a backbone with 8 linear layers each with ReLU non-linearity which takes the position, the expression and latent code as input (concatenated as vector $I$).
    	The output of the backbone is used to compute the density $\sigma$.
    	To compute the color, the output of the backbone is concatenated with the encoded viewing direction and inputted into another 4 linear layers with ReLU activations.
}	
		\label{fig:arch}
	\end{center}    
	\vspace{0.5cm}
}]

\section{Network Architecture}

We provide additional details of the proposed dynamic neural radiance fields architecture. As mentioned in the main paper, the dynamic neural radiance field is represented as a multi-layer perceptron (MLP).
In Fig.~\ref{fig:arch}, we depict the underlying architecture.

The dynamic neural radiance field is controlled by the expression coefficients that correspond to the blendshape basis of the used face tracker~\cite{thies2016face}.
To compensate for missing information, we also feed in the learned latent codes $\gamma$.
For a given sample location $(x,y,z)$ and the corresponding viewing direction $\vec{d}$, the MLP outputs the color and density which is used for the volumetric rendering, explained in the main document.
The MLP is based on a backbone of $8$ fully-connected layers, each $256$ neurons-wide, followed by ReLu as activation functions.
These activations are fed through a single layer to predict the density value, as well as a $4$-layer, $128$ neuron-wide branch to predict the final color value of the query point.